\documentclass{bmvc2k}
\usepackage{amssymb}
\usepackage{booktabs}
\usepackage{ulem}
\usepackage{amsfonts}
\usepackage{bbm}
\usepackage{float}

\title{MonoGS++: Fast and Accurate Monocular RGB Gaussian SLAM}

\addauthor{Ren-Wu Li}{renwu.li@amd.com}{1}
\addauthor{Wenjing Ke}{wenjing.ke@amd.com}{1}
\addauthor{Dong Li}{d.li@amd.com}{1}
\addauthor{Lu Tian}{lu.tian@amd.com}{1}
\addauthor{Emad Barsoum}{emad.barsoum@amd.com}{1}

\addinstitution{
 Advanced Micro Devices, Inc. \\
 Beijing, China
}

\runninghead{LI, ET AL}{MONOGS++: FAST AND ACCURATE MONOCULAR RGB GAUSSIAN SLAM}

\begin{document}

\maketitle

\begin{abstract}
We present MonoGS++, a novel fast and accurate Simultaneous Localization and Mapping (SLAM) method that leverages 3D Gaussian representations and operates solely on RGB inputs. 
While previous 3D Gaussian Splatting (GS)-based methods largely depended on depth sensors, our approach reduces the hardware dependency and only requires RGB input, leveraging online visual odometry (VO) to generate sparse point clouds in real-time. To reduce redundancy and enhance the quality of 3D scene reconstruction, we implemented a series of methodological enhancements in 3D Gaussian mapping. Firstly, we introduced dynamic 3D Gaussian insertion to avoid adding redundant Gaussians in previously well-reconstructed areas. Secondly, we introduced clarity-enhancing Gaussian densification module and planar regularization to handle texture-less areas and flat surfaces better. We achieved precise camera tracking results both on the synthetic Replica and real-world TUM-RGBD datasets, comparable to those of the state-of-the-art. Additionally, our method realized a significant 5.57x improvement in frames per second (fps) over the previous state-of-the-art, MonoGS~\cite{monogs}. 
\end{abstract}
\section{Introduction}
\label{sec:intro}

\paragraph{}
Simultaneous Localization and Mapping (SLAM) technologies play a pivotal role in the realm of robotics and augmented reality.
Conventional visual SLAM systems often face challenges with sparse and incomplete scene reconstructions, which are limited by the use of sparse point clouds. This limitation has propelled the development of methods like Neural Radiance Fields (NeRF), which facilitate dense and continuous scene reconstructions. However, the high computational load and the latency in scene updates inherent in NeRF-based systems often offset their benefits.

\begin{figure*}[ht]
    \centering
    \includegraphics[width=\textwidth]{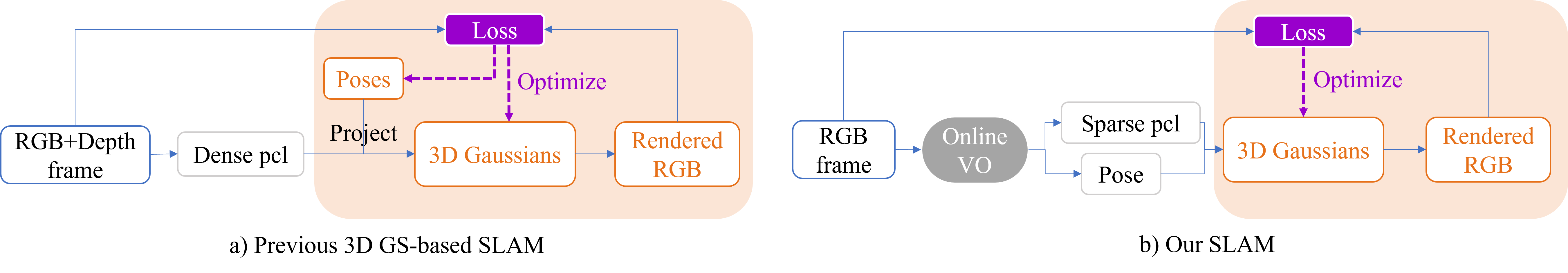}
    \caption{ The pipeline of previous 3D GS-based methods and our method. }
    \label{fig:fig_teaser}
\end{figure*}

A promising alternative to NeRF, 3D Gaussian Splatting (3D GS~\cite{gaussian_splatting}), offers a flexible, point-based scene representation through the use of 3D Gaussian distributions. This technique has been integrated into recent neural dense visual slam system such as SplaTAM~\cite{splatam} and Gaussian Splatting SLAM (MonoGS~\cite{monogs}), which have shown significant improvements both in speed and mapping capabilities compared to NeRF-based SLAM. 
However, these previous 3D GS-based approaches still face two main limitations. First, since the original 3D GS is initialized by the offline reconstructed structure from motion (SFM) point cloud, how to initialize the 3D Gaussian map is challenging for an online SLAM system. As depicted in Figure~\ref{fig:fig_teaser} a), previously mentioned 3D GS-based systems heavily rely on RGB-D images as input, from which local point clouds are back-projected and by the way local 3D Gaussians are initialized from current viewpoint. The dependency on depth sensor considerably narrows their applicability.
Second, to obtain the online camera poses for merging local 3D Gaussians, the existing 3D GS-based SLAM methods often integrate pose optimization into overall optimization process as shown in Figure~\ref{fig:fig_teaser} a), which burdens the system and slows down the overall latency.
% Second, to construct a global consistent 3D Gaussian map merged from local 3D Gaussians, camera poses must be estimated in an online fashion to transform local 3D Gaussians from camera coordinate system to world coordinate system. The existing 3D GS-based SLAM methods often suffer from slow and imprecise camera pose estimation due to their reliance on iterative optimization processes that handle extensive 3D Gaussians as show in Figure~\ref{fig:fig_teaser} a). Such computational intensity not only slows down the tracking but also affects the overall system latency, which is critical for real-time applications.

% Addressing these challenges, our approach aims to overcome the limitations of existing methods by pioneering a novel approach to initialize 3D GS with pure RGB data, enhancing the versatility and applicability of SLAM systems. Our proposed system leverages a differentiable, sparse patch-based visual odometry as the tracking frontend, which is not only capable of delivering real-time and accurate camera poses but also generates point clouds from monocular RGB inputs. As the system processes subsequent frames, the 3D Gaussian map incrementally evolves by integrating optimized patches from our visual odometry frontend.

% As depicted in Figure~\ref{fig:fig_teaser}, previous 3D GS-based methods relied on RGB-D input to generate dense point clouds and optimize the camera poses in the optimization, whereas our pipeline requires only RGB images to perform online generation of subsequent 3D Gaussians.
To address the above two limitations, we introduced MonoGS++, a monocular RGB Gaussian SLAM taking pure monocular RGB as input, as shown in Figure~\ref{fig:fig_teaser} b). We separated pose optimization from the overall network optimization, using visual odometry (VO) to derive online sparse point clouds and camera poses, and focused on enhancing the quality of 3D Gaussian mapping. With this design improvement and innovative enhancements in the mapping module, we developed a fast and accurate monocular RGB 3D Gaussian SLAM system.
% Our system achieves real-time tracking by a differential sparse patch-based visual odometry and incremental mapping of environments by optimizing 3D Gaussians, while significantly reducing the number of optimizable parameters to enhance efficiency. This reduction is achieved by avoiding the introduction of redundant Gaussians near well-optimized ones, thus streamlining the mapping process. Our approach broadens the potential applications of neural dense SLAM, making it feasible in diverse environments where only RGB inputs are available.

Our main contributions can be summarized as follows. (1) We proposed dynamic 3D Gaussian insertion to avoid adding redundant 3D Gaussians in previously well-reconstructed areas. (2) We proposed the clarity-enhancing Gaussian densification module and planar regularization to better handle texture-less areas and flat surfaces. (3) We achieved precise camera tracking results both on the synthetic Replica and real-world TUM-RGBD datasets, comparable to those of the state-of-the-art. Our tracking accuracy and rendering quality on the TUM-RGBD dataset surpass those of MonoGS~\cite{monogs}, with improvements of +5.21dB in PSNR, +0.18 in SSIM, and a reduction of 28.85cm in ATE. Furthermore, our method achieves a 5.57x increase in fps compared to MonoGS.

\section{Related Works}
\label{sec:review}

\paragraph{Implicit Neural Scene Representation.} Neural Radiance Fields (NeRF)\cite{nerf} and its follow-ups have transformed 3D scene representation by encoding scenes as continuous implicit fields using MLPs. These techniques excel in novel view synthesis and reconstruction, adept at capturing intricate geometries and unseen regions. The NeRF-based SLAM approach iMap\cite{sucar2021imap} integrates NeRF into SLAM, jointly optimizing camera poses and the implicit MLP. However, vanilla NeRF~\cite{nerf} requires excessive training time and is prone to overfitting. NICE-SLAM~\cite{zhu2022nice} and NICER-SLAM~\cite{zhu2023nicer} address this by using feature-dense grids and a pretrained encoder, achieving faster convergence and more accurate reconstructions. Other methods~\cite{voxfusion, eslam} accelerate training and rendering with explicit 3D voxels or hash grids, but rely on RGB-D data.

\paragraph{Point-based Neural Dense Visual SLAM.} Point-based scene representations have shown potential in neural dense visual SLAM. Point-SLAM~\cite{pointslam} employs a neural point cloud for concurrent tracking and mapping, ensuring accurate 3D reconstructions. 3D Gaussians Splatting (3D GS)\cite{gaussian_splatting}, a general point-based representation, is utilized in works like SplaTAM\cite{splatam}, and MonoGS~\cite{monogs}, delivering impressive results in tracking and mapping. Nonetheless, these methods often require depth data for initializing 3D Gaussians and are incompatible with monocular setups. While neural rendering in 3D Gaussian Splatting is fast, the iterative optimization in these SLAM frameworks can still slow down the overall performance.
\begin{figure*}[ht]
    \centering
    \includegraphics[width=1.0\textwidth]{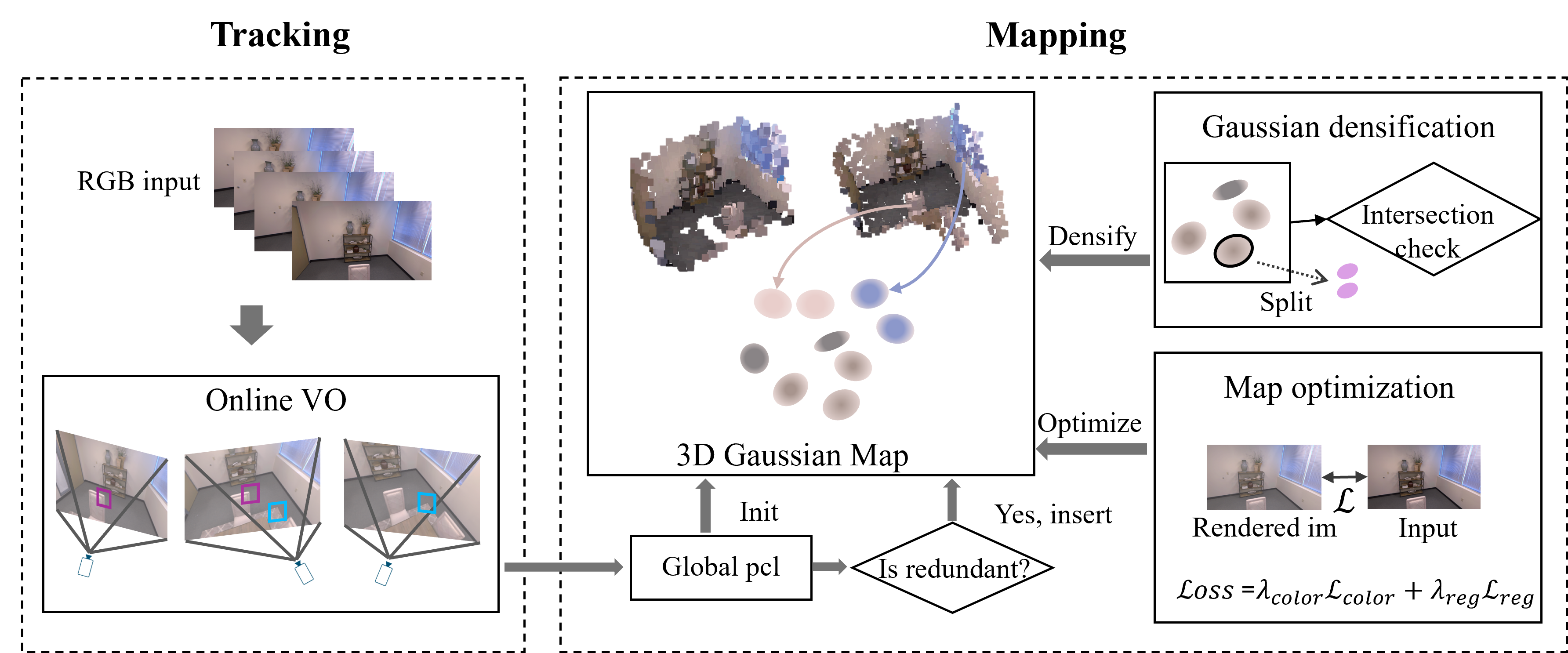}
    \caption{ Overview of the SLAM system.}
    % Our proposed system leverages a differentiable, sparse patch-based visual odometry as the tracking frontend, which is not only capable of delivering real-time and accurate camera poses but also generates point clouds from monocular RGB inputs. As the system processes subsequent frames, the 3D Gaussian map incrementally and is optimized by minimize the loss between rendered image and monocular RGB input.
    \label{fig:fig_pipeline}
\end{figure*}
\section{Methodology}
\label{sec:method}

\paragraph{}
As shown in Figure~\ref{fig:fig_pipeline}, our SLAM system mainly includes two parts: tracking and mapping. In tracking phase, we employ a deep patch-based visual odometry that is both efficient and lightweight, facilitating the estimation of camera poses alongside a set of optimized 3D sparse patches. In mapping phase, the scene is represented as a set of 3D Gaussians, which grows progressively and dynamically.

\subsection{Scene Representation}
\paragraph{}
Our system models the scene using a set of 3D Gaussians, each characterized by a mean vector $\mu \in \mathbb{R}^3$, which denotes position, and a covariance matrix $\Sigma$, defining its spatial distribution:
\begin{equation}
G(x) = \exp\left(-\frac{1}{2}(x-\mu)^T\Sigma^{-1}(x-\mu)\right), \quad \Sigma = RSS^TR^T
\end{equation}
Gaussians are further parameterized by color coefficients $c \in \mathbb{R}^k$, orientation via quaternion $r \in \mathbb{R}^4$, scale vector $s = \text{diag}(S) \in \mathbb{R}^3 $, and opacity $\alpha \in \mathbb{R}$.

In the Gaussian Splatting process, these 3D Gaussians are projected onto the image plane, transitioning to 2D representations. The 2D covariance matrix $\Sigma^{2D}$ is calculated as:
\begin{equation}
\Sigma^{2D} = JW\Sigma W^TJ^T
\end{equation}
where $J$ is the Jacobian matrix, and $W$ denotes the viewing transformation matrix.

The image is then synthesized by alpha-blending the colors and opacities of intersected Gaussians at each pixel $\mathbf{p} = (u, v)$:
\begin{equation}
C(\mathbf{p}) = \sum_{i=1}^{N} c_i(\mathbf{p}) \alpha_i(\mathbf{p}) \prod_{j=1}^{i-1} (1 - \alpha_j(\mathbf{p}))
\end{equation}
Here, $c_i(\mathbf{p})$ and $\alpha_i(\mathbf{p})$ represent the color and opacity of each Gaussian intersected by the ray from pixel $\mathbf{p}$, respectively.

\subsection{SLAM}

\subsubsection{Tracking}
\paragraph{}
We utilize Deep Patch Visual Odometry (DPVO~\cite{dpvo}), a learning-based, sparse monocular odometry method inspired by DSO~\cite{dso}. For each RGB keyframe $I_{i}$, DPVO samples $K$ square patches of size $p$, each parameterized in homogeneous coordinates as $P_{k}^{i} = [u, v, 1, d]^{\top}$, where $d$ represents the inverse depth, assumed constant across the patch.

DPVO constructs a patch graph with edges $\mathcal{E} = {(i, j, k)}$, where each edge indicates a trajectory of patch $P_{k}^{i}$ from $I_{i}$ to $I_{j}$. $I_{j}$ is within the local optimization window $\mathcal{N}(i)$. The projection of $P_{k}^{i}$ to $I_{j}$, denoted as $P_{k}^{j \leftarrow i}$, is derived from:
\begin{equation}
P_{k}^{j \leftarrow i} \sim K T_{j} T_{i}^{-1} K^{-1} P_{k}^{i},
\end{equation}
Camera poses and inverse depths are refined by differentiable bundle adjustment. A recurrent network predicts patch trajectory updates $\delta_{kj} \in \mathbb{R}^2$ and confidence weights $\Sigma_{kj} \in \mathbb{R}^2$ for each edge $(i, j, k)$ in the graph, aiming to minimize the Mahalanobis distance:
\begin{equation}
\sum_{(i, j, k) \in \mathcal{E}} \left\Vert K T_{j} T_{i}^{-1} K^{-1} \hat{P}{k}^{i} - \hat{P}{k}^{j \leftarrow i}\right\Vert_{\Sigma_{kj}}^2,
\end{equation}
For each new coming keyframe, the tracking module will optimize current camera pose and update the inverse depths of available patches.

\subsubsection{Mapping}

\paragraph{Map Initialization.}
The initialization of the 3D Gaussian map, denoted as $\mathcal{G}$, commences following the completion of the differential visual odometry's initialization stage, which spans $N$ frames. Contrary to traditional methods that employ depth sensors to back-project depth maps for initializing 3D Gaussians, our approach utilizes the centers of optimized patches from the initial $N$ frames. These centers are back-projected to form a global point cloud $\mathcal{P}$ in the world coordinate system, as described by the following equation:
\begin{equation}
\mathcal{P} = \{ T_{i}^{-1} K^{-1} \hat{P}_{k}^{i} \mid i \leq N, k \leq K \}.
\end{equation}

Subsequently, the map $\mathcal{G}$ is generated from $\mathcal{P}$, with the total number of 3D Gaussians given by $|\mathcal{G}| = N \times K$.

\paragraph{Dynamic 3D Gaussian Insertion.}
The Gaussian map $\mathcal{G}$ undergoes progressive enlargement by integrating each new keyframe $I_j$ along with its optimized camera pose $T_j$, as derived from the tracking module. This integration process involves back-projecting patches to update and augment the existing point cloud $\mathcal{P}$ by the updated point cloud $\mathcal{P}^{'} =  \{ T_{i}^{-1} K^{-1} \hat{P}_{k}^{i} \mid i \leq j, k \leq K \}$. Instead of generating a new 3D Gaussian for every point in the updated point cloud $\mathcal{P}'$, the necessity of each point is evaluated. Points situated within well-reconstructed regions are considered redundant and excluded. The inclusion of a point as the center of a new 3D Gaussian is determined by measuring the distance between each point in $\mathcal{P}'$ and the means of existing 3D Gaussians. Points with distances surpassing a threshold $\tau$ are selected for the creation of new 3D Gaussians:
\begin{equation}
\mathcal{P} = \mathcal{P} \cup \{ q \mid d(q, \mathcal{P}) < \tau, q \in \mathcal{P}^{'} \}, \quad
d(q, \mathcal{P}) = \min_{p \in \mathcal{P}} \Vert q - p \Vert_2.
\end{equation}

\paragraph{Clarity-Enhancing Gaussian Densification.}
In the original 3DGS~\cite{gaussian_splatting}, the 3D Gaussians are densified with the guidance of pixel rendering gradient, 3D Gaussians with high gradients are cloned or split depending on their 3D scales. As discussed in ~\cite{minisplatting}, the gradient-based densification may fail in areas with smooth textures and the rendered pixel is dominated by the 3D Gaussian with the largest alpha blending weight. Therefore, we further perform 3D Gaussian densification guided by rendering the most dominated Gaussian of each pixel to enhance the clarity of smooth colored regions. 
Specifically, for each pixel $\mathbf{p} = (u, v)$ in the current frame $I_{i}$, let $\mathcal{G}_{\mathbf{m}}(\mathbf{p})$ denote the 3D Gaussian with the largest alpha blending weight: 
\begin{equation}
    w_{m} = \alpha_m(\mathbf{p}) \prod_{n=1}^{m-1} (1 - \alpha_n(\mathbf{p})), \quad \mathbf{m} = \max_{m} w_{m}.
\end{equation}
Subsequently, we compute the intersected rendered pixels between $\mathcal{O}_{i}(n)$ and $\mathcal{O}_{i}(\mathbf{m})$ to produce a split mask $\mathbb{M}$:
\begin{equation}
\mathcal{M}(n) = \mathbbm{1}( |\mathcal{O}_i(n) \cap \mathcal{O}i(\mathbf{m})| > \sigma),
\end{equation}
where $\sigma$ is the split threshold. Gaussians $\mathcal{G}_{n}(\mathbf{p})$ for which $\mathcal{M}(n) = 1$ are then split to enhance the representation's fidelity and clarity in texture-smooth areas. 

\paragraph{Map Optimization with Planar Regularization.}
The Gaussian map is optimized by minimizing the discrepancy between the rendered image $\hat{I}_i$ and the original $I_i$. According to GaussianShader~\cite{gaussianshader} and GaussianPro~\cite{gaussianpro}, 3D Gaussians naturally tend to flatten and approximate planar surfaces during optimization. This characteristic is particularly advantageous for representing thin and flat structures such as walls, tables, and floors. Consequently, beyond the standard photometric loss $\mathcal{L}{color}$, we have integrated a planar regularization term $\mathcal{L}{reg}$. This term promotes the flattening of 3D Gaussian planes by specifically minimizing the smallest scale dimension across the three axes. And the objective functions is defined as:
\begin{equation}
\mathcal{L}_{color} = (1 - \lambda_{photo}) \cdot \mathcal{L}_{photo}(\hat{I}_i, I_i) + \lambda_{photo} \cdot \mathcal{L}_{SSIM}(\hat{I}_i, I_i), \quad \mathcal{L}_{reg} = \left| \max(0.01, \min(s)) \right|
\end{equation}
\begin{equation}
\mathcal{L} = \lambda_{color} \cdot \mathcal{L}_{color} + \lambda_{reg} \cdot \mathcal{L}_{reg}
\end{equation}
where $\mathcal{L}_{photo}$ denotes the photometric loss (i.e., the L1 loss), and $\mathcal{L}_{SSIM}$ signifies the structural similarity index measure. The weighting parameter $\lambda_{photo}$ is empirically set to 0.2, consistent with the Gaussian Splatting approach as delineated in~\cite{gaussian_splatting}.

\section{Experiments}
\label{sec:exp}

\subsection{Experimental Setting}
\paragraph{Datasets.}
We evaluate our method using both synthetic and real datasets, specifically Replica ~\cite{replica} and TUM-RGBD~\cite{tum-rgbd}. The Replica dataset presents fewer challenges in the RGB-D context due to its high-quality depth maps and minimal frame-to-frame displacement. However, it introduces significant challenges for monocular setups due to its textureless surfaces and purely rotational movements. In contrast, TUM-RGBD is a more challenging real-world dataset that poses more difficulties due to motion blur and noise resulting from the use of outdated, low-quality cameras. Additionally, the captured depth images are noisy and contain holes, which complicates processing with RGB-D based methods.

\paragraph{Metrics.}
We assess the performance of our method using RMSE of ATE for tracking accuracy and photometric metrics like PSNR, SSIM, and LPIPS for reconstruction quality. 
% These measures help quantify the accuracy, fidelity, and perceptual quality of the reconstructions against the actual scenes.

\paragraph{Baselines.}
We compare our approach with state-of-the-art neural dense visual slam methods, including NICE-SLAM~\cite{zhu2022nice}, Vox-Fusion~\cite{yang2022vox}, and Point-SLAM~\cite{pointslam}, as well as recent 3D GS-based methods such as SplaTAM~\cite{splatam} and MonoGS~\cite{monogs}. Notably, MonoGS~\cite{monogs}, which supports both monocular RGB and RGB-D modes, serves as our primary baseline for comparison. For a fair comparison, we reproduced the experiments for SplatAM, PointSLAM, and MonoGS and recorded the results in the experimental tables and visualization figures. All baseline methods rely exclusively on RGB-D data inputs, while only MonoGS (RGB)~\cite{monogs} and our method accept monocular RGB data as input.

\begin{table}[htbp]
\centering
\resizebox{0.95\textwidth}{!}{
    \begin{tabular}{cccccccccccc}
    \hline 
    Method  & Modality              & Metric               & room0 & room1 & room2 & office0 & office1 & office2 & office3 & office4 & Avg.  \\ \hline \hline
NICE-SLAM   & RGB-D          & PSNR{[}dB{]}$\uparrow$   & 22.12 & 22.47 & 24.52 & 29.07   & 30.34   & 19.66   & 22.23   & 24.94   & 24.42 \\
                   &   & SSIM$\uparrow$           & 0.68  & 0.75 & 0.81 & 0.87   & 0.88   & 0.79   & 0.80   & 0.85   & 0.80 \\
                &  & LPIPS$\downarrow$        & 0.33  & 0.27 & 0.20 & 0.22   & 0.18   & 0.23   & 0.20   & 0.19   & 0.233 \\
                 &     & ATE-MSE (cm)$\downarrow$ & 0.97  & 1.31  & 1.07  & 0.88    & 1.00    & 1.06    & 1.10    & 1.13    & 1.07  \\ \hline
Vox-Fusion  & RGB-D          & PSNR{[}dB{]}$\uparrow$   & 22.39 & 22.36 & 23.92 & 27.79   & 29.83   & 20.33   & 23.47   & 25.21   & 24.41 \\
                   &   & SSIM$\uparrow$           & 0.68  & 0.75 & 0.79 & 0.85   & 0.87   & 0.79   & 0.80   & 0.84   & 0.80 \\
                   &   & LPIPS$\downarrow$        & 0.30  & 0.26 & 0.23 & 0.24   & 0.18   & 0.24   & 0.21   & 0.19   & 0.236 \\
                   &   & ATE-MSE (cm)$\downarrow$ & 1.37  & 4.70  & 1.47  & 8.48    & 2.04    & 2.58    & 1.11    & 2.94    & 3.09  \\ \hline
Point-SLAM   & RGB-D         & PSNR{[}dB{]}$\uparrow$   & 32.40 & 34.08 & 35.5  & 38.26   & 39.16   & 33.39   & 33.48   & 33.49   & 35.17 \\
                    &  & SSIM$\uparrow$           & \underline{0.97}  & \underline{0.97} & \underline{0.98} & \underline{0.98}   & \underline{0.98}   & \underline{0.96}    & \textbf{0.96}   & \textbf{0.97}   & \underline{0.97} \\
                   &   & LPIPS$\downarrow$        & 0.11  & 0.11 & 0.11 & 0.1     & 0.11   & 0.15   & 0.13   & 0.14   & 0.12 \\
                   &   & ATE-MSE (cm)$\downarrow$ & 0.61  & 0.41  & 0.37  & \underline{0.38}    & 0.48    & 0.54    & 0.69    & 0.72    & 0.53  \\ \hline
SplaTAM     & RGB-D          & PSNR{[}dB{]}$\uparrow$   & 32.86  & 33.89 & 35.25 & 38.26   & 39.17   & 31.97   & 29.70   & 31.81   & 34.11 \\
               &       & SSIM$\uparrow$           & \textbf{0.98}  & \textbf{0.97}  & \textbf{0.98}  & \textbf{0.98}    & \textbf{0.98}    & \textbf{0.97}    & 0.95    & \underline{0.95}    & \textbf{0.97}  \\
               &       & LPIPS$\downarrow$        & \underline{0.07}  & 0.10  & 0.08  & 0.09    & 0.09    & 0.10    & 0.12    & 0.15    & 0.10  \\
               &       & ATE-MSE (cm)$\downarrow$ & \underline{0.31}  & \underline{0.40}  & \underline{0.29}  & 0.47    & \underline{0.27}    & \textbf{0.29}    & 0.32    & \underline{0.55}    & \underline{0.36}  \\ \hline
MonoGS      & RGB-D          & PSNR{[}dB{]}$\uparrow$   & \textbf{34.83} & \underline{36.43} & \textbf{37.49} & \underline{39.95}   & \underline{42.09}   & \textbf{36.24}   & \textbf{36.70}   & \underline{37.06}   & \underline{37.50} \\
                &      & SSIM$\uparrow$           & 0.95 & 0.95 & 0.96 & 0.97   & 0.97   & 0.96   & \underline{0.96}   & 0.95   & 0.96 \\
               &       & LPIPS$\downarrow$        & \textbf{0.068} & \textbf{0.076} & \textbf{0.075} & \underline{0.072}   & \textbf{0.055}   & \textbf{0.078}   & \textbf{0.065}   & \underline{0.099}   & \textbf{0.070} \\
               &       & ATE-MSE (cm)$\downarrow$ & 0.47  & 0.43  & 0.31  & 0.70    & 0.57    & \underline{0.31}    & \underline{0.31}    & 3.2     & 0.79  \\ \hline
MonoGS & RGB          & PSNR{[}dB{]}$\uparrow$   &28.94  &26.12  & 31.82 &  32.73  &34.47  &27.01  &30.76  &27.29  &29.89  \\
                &      & SSIM$\uparrow$          &0.88  &0.80  & 0.92 &  0.92  &0.93    &0.88    &0.91    &0.90  & 0.89 \\
               &       & LPIPS$\downarrow$        &0.18  &0.32  & 0.16 &  0.21  &0.19    &0.26    &0.16    &0.25 & 0.22 \\
                &      & ATE-MSE (cm)$\downarrow$ &5.87  &29.47  & 6.53 &  23.02  &15.93    &20.89    &3.98  &43.85 & 18.69 \\ \hline
\textbf{Ours}  & RGB       & PSNR{[}dB{]}$\uparrow$   & \underline{33.75} & \textbf{36.47} & \underline{37.01} & \textbf{42.31}   & \textbf{43.05}   & \underline{36.11}   & \underline{36.34}   & \textbf{37.28}   & \textbf{37.79} \\
                &      & SSIM$\uparrow$           & 0.94  & 0.96 & 0.96 & 0.98   & 0.97   & 0.95   & 0.96   & 0.96   & 0.96 \\
               &       & LPIPS$\downarrow$        & 0.092  & \underline{0.076} & \underline{0.077} & \textbf{0.052}   & \underline{0.064}   & \underline{0.090}   & \underline{0.078}   & \textbf{0.086}   & \underline{0.077} \\
               &       & ATE-MSE (cm)$\downarrow$ & \textbf{0.20}   & \textbf{0.17}  & \textbf{0.22}  & \textbf{0.29}    & \textbf{0.13}    & 0.42    & \textbf{0.20}    & \textbf{0.42}    & \textbf{0.26}  \\ \hline
    \end{tabular}
}
\caption{Quantitative results and comparison of SLAM method metrics on Replica dataset. The best results are shown in \textbf{bold}, while the second-best results are \underline{underlined}. }
%By default, MonoGS takes RGB-D as input for Replica dataset, here we also report the results in RGB mode for MonoGS, donated as MonoGS$^lozenge$.
\label{tab:tab_replica}
\end{table}

\vspace{-1.5mm}
\subsection{Evaluation}

\begin{figure}[htbp]
\centering
\includegraphics[width=0.95\textwidth]{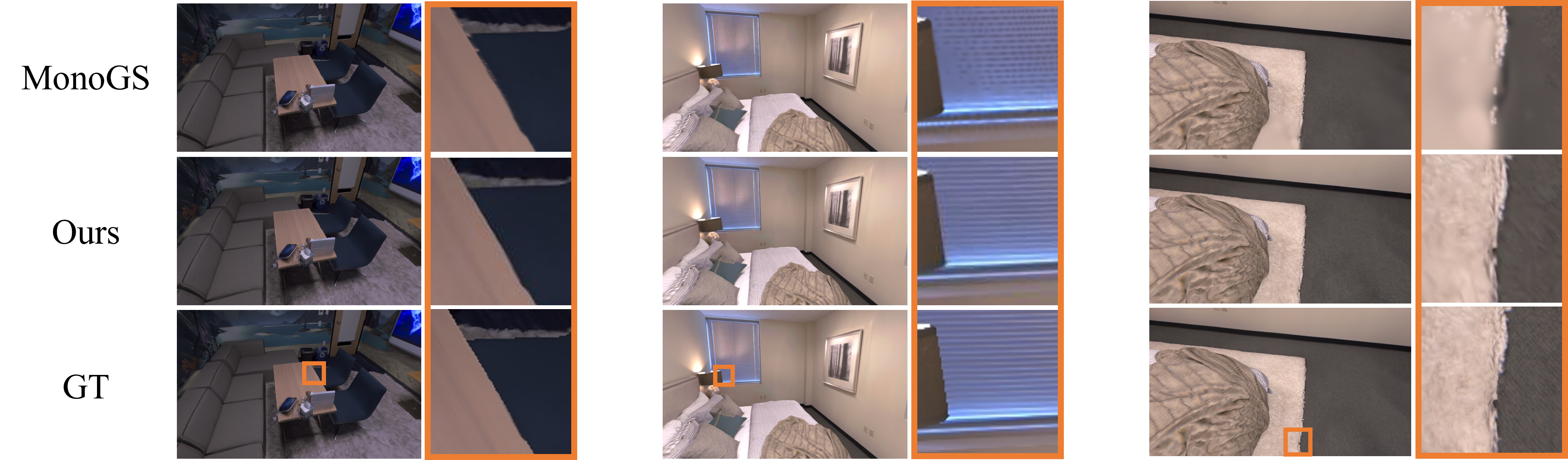}
\vspace{-2mm}
\caption{Rendering results on Replica dataset. Orange boxes highlight magnified details to emphasize quality differences.}
\label{fig:fig_replica}
\end{figure}

\paragraph{Results on Replica dataset.}
% Notably, recent implementations of MonoGS~\cite{monogs} (Gaussian Splatting SLAM) have reported difficulties in performing adequately on the Replica dataset, particularly under conditions of pure rotational camera motion. 
The results on Replica dataset are presented in Table~\ref{tab:tab_replica}. This table shows that our method not only achieves the most accurate camera tracking outcomes but also obtains comparable results in rendering quality. Our approach achieves better PSNR scores on average. Additionally, our method delivers competitive performance in SSIM and LPIPS. Compared to MonoGS (RGB), which uses the same RGB input as we do, our performance metrics significantly surpass theirs across all indicators. Since most methods achieve good accuracy, we focused on detailed comparisons with MonoGS, which has the highest accuracy among them. As shown in Figure~\ref{fig:fig_replica}, our method performed better on planar details like table edges and blinds due to effective planar regularization. In weakly textured areas such as carpets, MonoGS's results were blurry, whereas ours retained more detail. This demonstrates the efficacy of our clarity-enhancing Gaussian densification in splitting 3D Gaussians on such regions, enhancing texture detail reproduction.
\vspace{-6mm}

% This comprehensive superiority underscores the robustness and precision of our approach, even in challenging visual conditions. 

\begin{table}[htbp]
\centering
\resizebox{0.8\textwidth}{!}{
    \begin{tabular}{ccccccccc}
    \hline
    Method & Modality & Metric & fr1/desk & fr1/desk2 & fr1/room & fr2/xyz & fr3/office & Avg. 
    \\ \hline \hline
    NICE-SLAM & RGB-D & PSNR[dB]$\uparrow$ & 13.83 & 12.00 & 11.39 & 17.87 & 12.89 & 13.59 \\ 
     & & SSIM$\uparrow$ & 0.56 & 0.51 & 0.37 & 0.71 & 0.55 & 0.54 \\ 
     & & LPIPS$\downarrow$ & 0.48 & 0.52 & 0.62 & 0.34 & 0.49 & 0.49 \\
     & & ATE-MSE (cm)$\downarrow$ & 4.26 & \underline{4.99} & 34.49 & 31.73 & 3.87 & 15.87 \\ \hline
    Vox-Fusion & RGB-D & PSNR[dB]$\uparrow$ & 15.79 & 14.12 & 14.20 & 16.32 & 17.27 & 15.54 \\ 
     & & SSIM$\uparrow$ & 0.64 & 0.56 & 0.56 & 0.70 & 0.67 & 0.63 \\ 
     & & LPIPS$\downarrow$ & 0.52 & 0.54 & 0.55 & 0.43 & 0.45 & 0.50 \\
     & & ATE-MSE (cm)$\downarrow$ & 3.52 & 6.00 & \underline{19.53} & 1.49 & 26.01 & 11.31 \\ \hline
    Point-SLAM & RGB-D & PSNR[dB]$\uparrow$ & 13.87 & 14.12 & 14.16 & 17.56 & 18.43 & 15.63 \\ 
     & & SSIM$\uparrow$ & 0.62 & 0.59 & 0.64 & 0.70 & 0.75 & 0.66 \\ 
     & & LPIPS$\downarrow$ & 0.54 & 0.56 & 0.54 & 0.58 & 0.44 & 0.53 \\
     & & ATE-MSE (cm)$\downarrow$ & 4.34 & \textbf{4.54} & 30.92 & 1.31 & 3.48 & 8.92 \\ \hline
     
    SplaTAM & RGB-D & PSNR[dB]$\uparrow$ & \underline{21.16} & \underline{19.26} & \underline{18.73} & \underline{23.11} & 19.92 & \underline{20.44} \\
     & & SSIM$\uparrow$ & \textbf{0.87} & \textbf{0.80} & \textbf{0.78} & \textbf{0.90} & \underline{0.82} & \textbf{0.83}\\
     & & LPIPS$\downarrow$ & \underline{0.24} & \underline{0.33} & \underline{0.33} & \underline{0.21} & \underline{0.34} & \underline{0.29} \\
     & & ATE-MSE (cm)$\downarrow$ & 3.35 & 6.54 & \textbf{11.86} & \underline{1.34} & 5.41 & \textbf{5.70} \\ \hline
     
    MonoGS & RGB-D & PSNR[dB]$\uparrow$ & 9.99& 8.90& 8.95& 12.46& 15.95& 11.25 \\ 
     & & SSIM$\uparrow$ & 0.36 & 0.31 & 0.46 & 0.71 & 0.46 & 0.46 \\ 
     & & LPIPS$\downarrow$ & 0.70 & 0.71& 0.60& 0.30& 0.74& 0.61 \\
     & & ATE-MSE (cm)$\downarrow$ & 20.21& 90.92& 104.32& 1.47& 104.88& 64.36 \\ \hline
     
     MonoGS & RGB & PSNR[dB]$\uparrow$ & 17.31 &14.06 &14.76 & 22.06 &\underline{23.02} & 18.24  \\ 
     & & SSIM$\uparrow$ &0.65 &0.50 &0.52 &0.72 &0.78 &0.63  \\ 
     & & LPIPS$\downarrow$ &0.38 &0.62 &0.60 &0.27 &0.32 &0.43  \\
     & & ATE-MSE (cm)$\downarrow$ &\underline{3.05} &79.45 &84.78 &4.31 &\underline{1.85} &34.68 \\ \hline \hline
    \textbf{Ours} & RGB & PSNR[dB]$\uparrow$ & \textbf{22.85} & \textbf{20.64}& \textbf{22.16} & \textbf{26.52} & \textbf{25.08} & \textbf{23.45} \\
     & & SSIM$\uparrow$ & \underline{0.82}& \underline{0.77}& \underline{0.77} & \underline{0.86} & \textbf{0.85} & \underline{0.81} \\
     & & LPIPS$\downarrow$ & \textbf{0.20}& \textbf{0.29}& \textbf{0.31} & \textbf{0.13} & \textbf{0.18} & \textbf{0.22} \\
     & & ATE-MSE (cm)$\downarrow$ & \textbf{1.79}& 5.18& 21.44& \textbf{0.38}& \textbf{0.36} & \underline{5.83}\\ \hline
    \end{tabular}
}
\caption{Quantitative results and comparison of SLAM method metrics on TUM-RGBD dataset. The best results are shown in \textbf{bold}, while the second-best results are \underline{underlined}.}
\label{tab:tab_tum}
\end{table}

\begin{figure}[hbtp]
\centering
\includegraphics[width=0.9\textwidth]{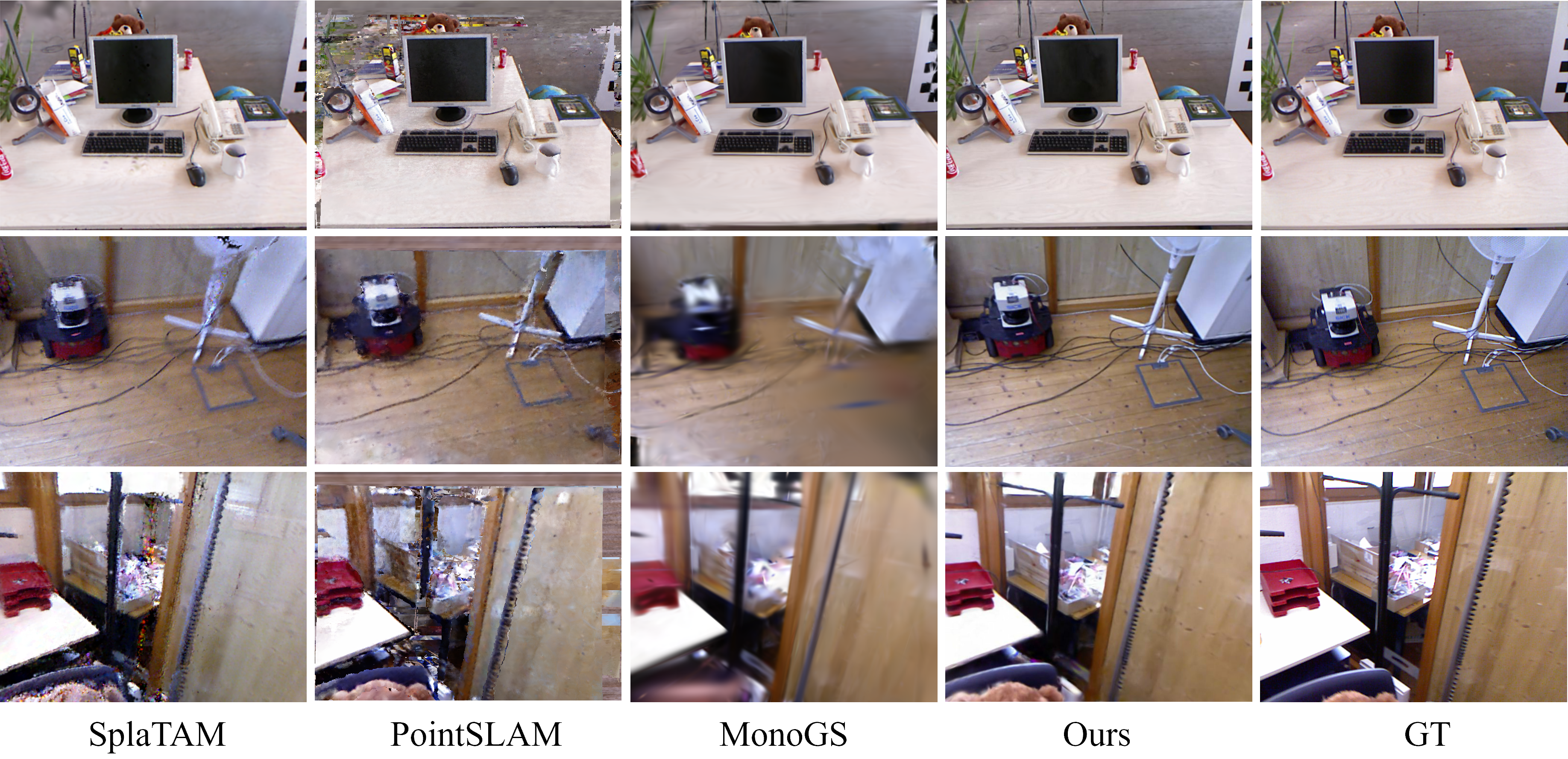}
\vspace{-4mm}
\caption{Rendering results on the TUM-RGBD dataset.}
\label{fig:fig_tum}
\end{figure}

\paragraph{Results on TUM-RGBD dataset.}
The results on Replica dataset are presented in Table~\ref{tab:tab_tum}. Our method achieves higher PSNR and LPIPS scores on all sequences and the second highest SSIM scores minimally lower than SplaTAM by 0.02. Regarding camera tracking accuracy, our method achieves better results on 3 sequences. However, it showed a significant discrepancy compared to SplaTAM on the fr1/room. Upon analysis, this scene involved a dynamically moving pedestrian and rapid camera motion, which led to instability in the tracking with purely RGB-based VO. It's worth noting that although MonoGS supports both RGB and RGB-D inputs, the experimental results indicate that it exhibits significant instability across different data qualities and input configurations. In Figure~\ref{fig:fig_tum}, SplaTAM displays imperfections along object edges like chessboards and metal rods. PointSLAM often show excessive noise, especially in small background objects. Meanwhile, MonoGS (RGB) generally results in blurry effects. In contrast, our method delivered superior rendering quality, excelling in weakly textured areas such as desktops and floors, and capturing fine details like the serrations on metal rods.

% Qualitative evaluation also performed on both Replica and TUM-RGBD datasets. As shown in Figure~\ref{fig:fig_replica}, we compare the rendered images with those rendered by MonoGS~\cite{monogs} (taking RGB-D as input). Our images exhibit fewer artifacts in thin areas such as windows and are less blurry in textureless regions like tables and the carpet. This improvement is largely attributed to our advanced Gaussian densification process, which effectively increases the presence of Gaussians in areas with smooth colors and flat structures, enhancing the overall image quality. For the TUM-RGBD dataset, as shown in Figure~\ref{fig:fig_tum}, we benchmarked against SplaTAM~\cite{splatam}, PointSLAM~\cite{pointslam}, and MonoGS~\cite{monogs}. 
% Despite the significant distortions and motion blur in the input images, which generally reduce rendering quality compared to the synthesized Replica dataset, our method remains robust and consistently delivers superior rendering quality, even with pure RGB input. This demonstrates our method's resilience and effectiveness in handling low-quality data and underscores its advantages in realistic SLAM applications.

\begin{table}[htbp]
\centering
\resizebox{0.75\textwidth}{!}{
\begin{tabular}{lccccc}
\hline
Method & Single Process & Tracking / Frame & Mapping / Frame & FPS $\uparrow$ & ATE $\downarrow$ \\
\hline
Point-SLAM &  & 8.20s & 40.80s & 0.060 & 0.61 \\
SplaTAM &  & 3.13s & 5.33s & 0.115 & 0.31 \\
MonoGS &  & 1.47s & 3.05s & 0.445 & 0.47 \\
MonoGS(RGB) &  & 3.40s & 11.60s & 0.152 & 5.87 \\
MonoGS* & $\checkmark$ & 0.74s & 15.62s & 0.246 & 0.39 \\
Ours & $\checkmark$ & \textbf{0.067s}& \textbf{0.27s}& \textbf{2.48}& \textbf{0.20} \\
\hline
\end{tabular}
}
\caption{Runtime analysis on Replica/Room0. All experimental results were obtained running on the same hardware configuration.}
\label{tab:tab_runtime}
\vspace{-4mm}
\end{table}
\subsection{Runtime and Efficiency}
\paragraph{}
In Table~\ref{tab:tab_runtime}, we present a comprehensive analysis of runtime statistics, comparing our method with several baselines. All experimental results were obtained running on the same hardware configuration. Among these baselines, SplaTAM, Point-SLAM and MonoGS all utilize multi-processing to enhance performance, whereas MonoGS* operates within a single process. 
% Among these baselines, SplaTAM-S represents a streamlined version of SplaTAM, which achieves efficiencies by reducing both the number of optimization iterations and the resolution of processed images. MonoGS utilizes multi-processing to enhance performance, whereas MonoGS* operates within a single process but incorporates a higher number of iterations. 
% We calculate Frames Per Second (FPS) by dividing the total frames processed by the total time consumed by each method. 
As shown in Table~\ref{tab:tab_runtime}, our method not only supports real-time camera tracking but also achieves the most accurate tracking results. Significantly, although our approach is implemented in a single process fashion the same as MonoGS*, our system achieves the highest FPS, operating 5.57 times faster than MonoGS, 16.32 times faster than MonoGS taking RGB as input and 10.08 times faster than MonoGS*. This performance demonstrates our approach's superior efficiency and prospects in real-time applications.

\begin{table}[htbp]
\centering
\resizebox{\textwidth}{!}{
\begin{tabular}{ccccccccc}
\hline
    & Dynamic Gaussian Insertion & Clarity-Enhancing Densification & Planar Regularization & PSNR$\uparrow$ & SSIM$\uparrow$ & LPIPS$\downarrow$ & \# of Gaussians & FPS$\uparrow$ \\ \hline
(A) &                            & $\checkmark$                    & $\checkmark$          & 36.85          & 0.963          & 0.080 & 1680278& 1.76\\
(B) & $\checkmark$               &                                 & $\checkmark$          & 35.88          & 0.955          & 0.110 & 769568& 2.71\\
(C) & $\checkmark$               & $\checkmark$                    &                       & \textbf{37.07}          & 0.963          & 0.083 & 1123479& 2.85\\ \hline
(D) & $\checkmark$               & $\checkmark$                    & $\checkmark$          & \underline{37.01}          & \textbf{0.963}          & \textbf{0.077} & 1143534& 2.84\\ \hline
\end{tabular}
}
\caption{Ablation study on Replica/Room2. We introduce four variants of our approach, designated as (A), (B), (C) and (D). A deterioration in accuracy is observed upon the removal of any proposed component.}
\vspace{-4mm}
\label{tab:tab_ablation}
\end{table}
\begin{figure}[htbp]
\centering
\includegraphics[width=0.75\textwidth]{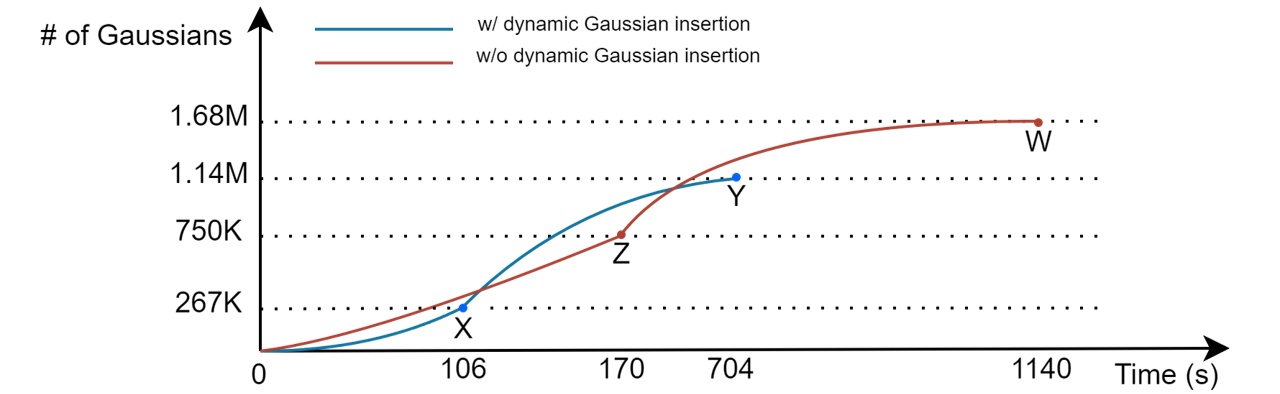}
\caption{Estimated number of Gaussians in the system over the time series. Variant (A) in \textcolor{red}{red} and (D) in \textcolor{blue}{blue}. }
\label{fig:fig_ablation}
\end{figure}
\subsection{Ablation Study}
\paragraph{}
To validate the effectiveness of each component of our method, we conducted experiments with four variations: (A) without the dynamic Gaussian insertion strategy; (B) without the clarity-enhancing Gaussian densification module; (C) without the planar regularization term; and (D) with all components enabled. The results validated on the Replica-Room2 sequence are presented in Table~\ref{tab:tab_ablation}. 
The dynamic Gaussian insertion primarily enhances efficiency, the clarity-enhancing Gaussian densification module significantly improves overall rendering quality, and the planar regularization term mainly boosts the LPIPS value.
% These results reveal that omitting the clarity-enhancing Gaussian densification module significantly deteriorates all measured metrics. Similarly, removing the planar regularization term notably reduces the LPIPS score, demonstrating the importance of each component in achieving optimal performance. 
% Further analysis from Table~\ref{tab:tab_ablation} shows that dynamic Gaussian insertion drastically reduces both the number of Gaussians and processing time. 
Besides, Figure~\ref{fig:fig_ablation} further analyzes the dynamic Gaussian insertion's impact.
% XY and ZW depict the map refinement stages post-tracking. 
With dynamic Gaussian insertion, the number of Gaussians prior to refinement is nearly one-third of that without it, facilitating faster optimization and achieving higher frame rates (2.84 fps vs. 1.76 fps).
\section{Conclusion}
\label{sec:conclusion}
We have proposed MonoGS++, a fast and accurate monocular RGB Gaussian SLAM taking pure RGB as input and performing 3D Gaussian mapping. This system achieves comparable accuracy with SOTA RGB-D-dependent methods and surpasses them on system efficiency by a large margin, promising broader applications in robotics and augmented reality. Future efforts will be focused on enhancing its robustness and adaptability in more challenging scenes with motion blur and dynamic objects moving.

\clearpage

\bibliography{egbib}
\end{document}